\documentclass[sigconf]{acmart}

\setcopyright{rightsretained}

\usepackage[utf8]{inputenc} 
\usepackage[T1]{fontenc}    
\usepackage{hyperref}       
\usepackage{url}            
\usepackage{booktabs}       
\usepackage{amsfonts}       
\usepackage{nicefrac}       
\usepackage{microtype}      
\usepackage{amsmath}
\usepackage{enumerate}
\usepackage{graphicx}
\usepackage{bm}
\usepackage{caption}
\usepackage{booktabs}      
\usepackage{amsfonts}    
\usepackage{nicefrac}      
\usepackage{comment}

\usepackage{grffile}
\usepackage{subcaption}
\usepackage[font=footnotesize,figurewithin=none]{caption}

\long\def\cut#1{}
\newcommand{\eg}{\textit{e.\,g., }}

\newcommand{\etal}{\textit{et al}.}

\newcommand{\curlyZ}{\mathcal{Z}}

\newcommand{\curlyL}{\mathcal{L}}
\newcommand{\xbf}{\bm{x}}
\newcommand{\boldeta}{\bm{\eta}}
\newcommand{\boldlambda}{\bm{\lambda}}
\newcommand{\boldtheta}{\bm{\theta}}
\newcommand{\boldthetahat}{\hat{\bm{\theta}}}

\newcommand{\boldphi}{\bm{\phi}}
\newcommand{\boldphihat}{\hat{\bm{\phi}}}

\newcommand{\ybf}{\bm{y}}
\newcommand{\zbf}{\bm{z}}
\newcommand{\expect}{\mathbb{E}}
\newcommand{\leda}{\mathcal{L}_{EDA}}
\newcommand{\footer}{\newline \small \it This is an expanded version of the paper by the same title published in the proceedings of the The Genetic and Evolutionary Computation Conference (GECCO), 2020}
\DeclareMathOperator*{\argmax}{argmax}
\DeclareMathOperator*{\argmin}{argmin}

\newcommand\blfootnote[1]{%
  \begingroup
  \renewcommand\thefootnote{}\footnote{#1}%
  \addtocounter{footnote}{-1}%
  \endgroup
}
 \acmConference[]{}{}{}
\usepackage{ifthen}
\newboolean{ARXIV}  
\setboolean{ARXIV}{True}

\newboolean{FOOTNOTEHEADER}
\setboolean{FOOTNOTEHEADER}{True}

\renewcommand\footnotetextcopyrightpermission[1]{} 
\begin{document}

\title[]{A view of Estimation of Distribution Algorithms through the lens of Expectation-Maximization}


\author{David Brookes}
\affiliation{%
  \institution{University of California, Berkeley}
  \city{Berkeley} 
  \state{CA}
  \postcode{94720}
}
\email{david.brookes@berkeley.edu}

\author{Akosua Busia}
\affiliation{%
  \institution{University of California, Berkeley}
  \city{Berkeley} 
  \state{CA} 
  \postcode{94720}
}
\email{akosua@berkeley.edu}

\author{Clara Fannjiang}
\affiliation{
  \institution{University of California, Berkeley}
    \city{Berkeley} 
  \state{CA} 
  \postcode{94720}
  }
\email{clarafy@berkeley.edu}

\author{Kevin Murphy}
\affiliation{%
  \institution{Google Research}
  \city{Mountain View}
  \state{CA}
  \postcode{94043}
}
\email{kpmurphy@google.com}

\author{Jennifer Listgarten}
\affiliation{%
  \institution{University of California, Berkeley}
    \city{Berkeley} 
  \state{CA} 
  \postcode{94720}
  }
\email{jennl@berkeley.edu}

\renewcommand{\shortauthors}{D. Brookes \textit{et. al.}}

\begin{abstract}
We show that a large class of Estimation of Distribution Algorithms, including, but not limited to, Covariance Matrix Adaption, can be written as a Monte Carlo Expectation-Maximization algorithm, and as exact EM in the limit of infinite samples. Because EM sits on a rigorous statistical foundation and has been thoroughly analyzed, this connection provides a new coherent framework with which to reason about EDAs.
\end{abstract}

%
%



\maketitle

\section{Introduction}

\ifthenelse{\boolean{FOOTNOTEHEADER}}{
\blfootnote{\footer}{}
}

Estimation of Distribution Algorithms (EDAs) are a widely used class of algorithms designed to solve optimization problems of the form $\zbf^* = \argmax_{\zbf\in\curlyZ} f(\zbf)$, where $f : \curlyZ \rightarrow \mathbb{R}$ is a function over a space of discrete or continuous inputs, $\curlyZ$. Instead of solving this objective directly, EDAs solve the related objective 

\begin{equation}\label{eq:mbo}
    \boldtheta^* = \argmax_{\boldtheta} \expect_{p(\zbf|\boldtheta)}[f(\zbf)],
\end{equation} where $p(\zbf|\boldtheta)$ is a probability density over $\curlyZ$, parameterized by $D$ parameters \mbox{$\boldtheta \in \mathbb{R}^D$}. When $p(\zbf|\boldtheta)$ has the capacity to represent point masses on the maxima of $f(\zbf)$, then these two formulations have the same optimal values. Reasons for using the latter formulation of Equation \ref{eq:mbo} include convenience for derivative-free optimization \cite{Hansen2001}, 
 enabling of rigorous analysis of associated optimization algorithms\citep{Zlochin2004}, and the leveraging of a probabilistic formulation to incorporate auxiliary information \citep{Brookes2019}. In many cases this objective is further modified by applying a monotonic {\it shaping} function, $W(\cdot)$ to $f(\zbf)$ \citep{Wierstra2008, Wierstra2014, Yi2009}, which alters convergence properties of algorithms to solve it, but does not change the optima. 

The general algorithmic template for EDAs is as follows. Beginning with an initial parameter setting, $\boldtheta^{(0)}$, of the parametrized density, $p(\zbf|\boldtheta)$, each iteration $t \in \{0,1,...T\}$ of an EDA generally consists of three steps:

\begin{enumerate}
    \item Draw $N$ samples, $\{\zbf_i\}_{i=1}^N$, from $p(\zbf|\boldtheta^{(t)})$.
    \item Evaluate $W(f(\zbf_i))$ for each $\zbf_i$.
    \item Find a $\boldtheta^{(t+1)}$ that uses the weighted samples and corresponding function evaluations to move $p(\zbf|\boldtheta)$ towards regions of $\curlyZ$ that have large function values.
\end{enumerate}

The last step is generally accomplished by attempting to solve
\begin{equation}\label{eq:eda_obj}
  \mbox{$\boldtheta^{(t+1)} = \argmax_{\boldtheta} \sum_{i=1}^N W(f(\zbf_i))\log p(\zbf_i|\boldtheta)$},
\end{equation}
which can be seen as a weighted maximum likelihood problem with weights $W(f(\zbf_i))$. We refer to this set of steps as a ``core'' EDA because it ties together most EDAs.

In the case of Covariance Matrix Adaptation (CMA-ES), this core algorithm is often modified in a variety of ways to improve performance. For example, samples from previous iterations may be used, directly or indirectly, in the last step,  resulting in smoothing of the parameter estimates across iterations. Adaptive setting of  parameter-specific step sizes, and ``path evolution'' heuristics \citep{Hansen2001,Ollivier2017, Brookes2019} are also common. These layers on top of the core EDA have generally been derived in a manner specific only to CMA-ES, and are not readily generalizable to other EDAs. For this reason, we restrict ourselves to the core EDA  algorithm just described. However, as we shall see, our formalism also allows for a large class of EDA parameter smoothing variations which encompass many, but not all, variations of CMA-ES.

EDAs are also distinguished by the choice of parametrized density, $p(\zbf|\boldtheta)$, which we refer to here as the ``search model''.\footnote{This object is often simply referred to as the "probability distribution" \cite{Kern2004}, however; we use "search model" to distinguish it from other probability distributions used herein.}  Many EDAs use exponential family models for the search model, most commonly the multivariate Gaussian as in CMA-ES. However, \\ Bayesian networks \citep{Ocenasek2002}, Boltzmann machines \citep{Hu2011,Shim2013}, Markov Random Fields \citep{Shakya2007}, Variational Autoencoder \citep{Brookes2019}, and others \citep{Kern2004} have also been used. Unless otherwise specified, our analysis in the following is quite general and makes no assumptions on the choice of search model.
\\
\\
\noindent {\bf Our contributions} 
We show that a large class of EDAs---including, but not limited to, variants of CMA-ES---can be written as Monte Carlo (MC) Expectation-Maximization (EM) \citep{Beal1998}, and in the limit of infinite samples, as exact EM \citep{Dempster77}. 
Because EM sits on a rigorous statistical foundation and has been thoroughly analyzed, this connection provides a new framework with which to reason about EDAs. Within this framework, we also show how many parameter-smoothed variants of CMA-ES can be written as Monte Carlo maximum {\it a posteriori}-EM (MAP-EM), which suggests a possible avenue to develop similar smoothing algorithms for EDAs with other search models. 
Finally, leveraging our EDA-EM connection, we provide a new perspective on how EDAs can be seen as performing approximate natural gradient descent.

\subsection{Related work}
 The Information Geometry Optimization (IGO) framework \citep{Ollivier2017} unifies a large class of approaches---including EDAs such as CEM \citep{Rubinstein1999} and CMA-ES \citep{Hansen2001}---for solving the objective in Equation \ref{eq:mbo} by discretizing a natural gradient flow. Instantiations of IGO are most readily seen as tantamount to using the score function estimator  (sometimes referred to as the ``log derivative trick'') on Equation \ref{eq:mbo}, combined with natural gradient, as in Natural Evolution Strategies \citep{Wierstra2008, Wierstra2014}.  The IGO framework does not connect to EM; therefore, IGO provides a complementary viewpoint to that presented herein.  

\citet{akimoto2012theoretical} show how \mbox{CMA-ES} with rank-$\mu$ updates, but without global step size and evolution paths, can be viewed as a natural evolution strategy. In this context, they briefly remark on how a simplified CMA-ES can be viewed as performing generalized EM with a partial-M step using a natural gradient. The technical underpinnings of this result are restricted specifically to CMA-ES because of the dependence on a Gaussian search model,  and do not readily generalize to the broader EDA setting considered herein.

Staines \etal~introduce the notion of variational optimization, by explicitly considering the fact that the optimum of the objective function in Equation \ref{eq:mbo} is a lower bound on the optimum of $f$. They also clearly delineate conditions under which the bound can be satiated, and the objective function is convex and has derivatives that exist \citep{Staines2013}; no connections to EDAs are made.

\section{Background: Free Energy view of EM} \label{sec:background}
\ifthenelse{\boolean{FOOTNOTEHEADER}}{
\blfootnote{\footer}{}
}

Before deriving the connection between EM and EDAs, we first review some needed background on EM. EM is a well-known method for performing maximum likelihood parameter estimation in latent variable models that exploits the structure of the joint likelihood between latent and observed variables \cite{Dempster77}. Intuitively, each E-step imputes the latent variables, and the subsequent M-step then uses these ``filled in'' data to do standard maximum likelihood estimation, typically in closed form. EM iterates between these E- and M- steps until convergence. We use the Free Energy interpretation of EM and its accompanying generalizations \citep{Neal1999} in order to reveal the connection between EDAs and EM. 

Let $\xbf$ and $\zbf$ be observed and latent variables, respectively. The task of maximum likelihood estimation in a latent variable model is to find $\boldphihat = \argmax_{\boldphi} \mathcal{L}(\boldphi)$ where
\begin{equation}
    \label{eq:lik}
    \mathcal{L}(\boldphi) = \log p(\xbf|\boldphi) =
   \log \int p(\xbf, \zbf|\boldphi) d\zbf,
\end{equation} 
for some model density $p$ parameterized by $\boldphi$.
In \citet{Neal1999}, the authors define a function known as the free energy, given by
\begin{equation}\label{eq:free_energy_def}
    F(q, \boldphi) = \mathbb{E}_{q(\zbf)}[\log p(\xbf, \zbf|\boldphi)] + H[q],
\end{equation} 
where $q(\zbf)$ is any probability density over the latent variables,  $H[q]$ is the entropy of $q$, and the term preceding the entropy is known as the {\it expected complete log-likelihood}. The free energy lower bounds the log-likelihood, $\mathcal{L}(\boldphi) \geq F(q, \boldphi)$, and this bound is satiated only when $q(\zbf)$ is equal to the true posterior, $q(\zbf)=p(\zbf|\xbf, \boldphi)$. If the true posterior cannot be calculated exactly, one may approximate it in one of several ways, two of which are described next.

In the first, the posterior is approximated by restricting it to a parameterized family of distributions, $q(\zbf|\bm{\psi})$, where both the parameters of the likelihood and the variational family must now be estimated. This is known as variational EM. Unless the true posterior lies in the parameterized class, which is not typically the case, the bound $\mathcal{L}(\boldphi) \geq F(q, \boldphi)$ cannot be satiated, leading to a {\it variational gap} given by $D_{KL}(q(\zbf)||p(\zbf|\xbf, \boldphi))$, where $D_{KL}$ denotes the KL divergence.

In another approximation, one draws samples from the true posterior, \mbox{ $\zbf_i \sim p(\zbf|\xbf, \boldphi^{(t)})$}, to estimate the expected complete log likelihood. This is known as Monte Carlo EM (MC-EM).
\cut{
, and can be seen as approximating the posterior with a mixture of weighted particles,
\mbox{ $q(\zbf) = \frac{1}{N}\sum_i \delta_{\zbf_i}(\zbf)$}.} This also  induces a ``gap'' between the true and approximate posterior (because of the use of finitely many samples), which can be similarly computed as $D_{KL}(q(\zbf)||p(\zbf|\xbf, \boldphi))$, and hence we will refer to it also as a variational gap.
One major distinction between variational and MC-EM is that variational EM explicitly attempts to minimize the variational gap at each iteration through optimization, while MC-EM only implicitly closes the gap as the number of samples goes to infinity. 

All of EM, MC-EM, and variational EM can be viewed as alternating coordinate descent on the free energy function ~\citep{Neal1999}. In particular, the alternating updates at iteration $t$ are given by: 
\begin{itemize}

    \item E-step: $q^{(t+1)}\gets \argmax_{q} F(q, \boldphi^{(t)})$, that is, compute or estimate the posterior over the hidden variables. \cut{Intuitively, this steps imputes  the ``missing'' data, $\{\zbf_i\}$.}
    It can be shown that this is equivalent to minimizing the variational gap by solving \mbox{
    $q^{(t+1)}(\zbf) = \argmin_{q} D_{KL}(q(\zbf)||p(\zbf|\xbf, \boldphi^{(t)})$}. 
    \item M-step:~\mbox{$\boldphi^{(t+1)}\gets \argmax_{\boldphi} F(q^{(t+1)}, \boldphi)$}, which is equivalent to maximizing the expected complete log likelihood with respect to $\boldphi$, \mbox{$\boldphi^{(t+1)} \gets \argmax_{\boldphi} \mathbb{E}_{q^{(t+1)}(\zbf)}[\log p(\xbf, \zbf|\boldphi)]$}. 
    \cut{This can also be seen as solving a weighted maximum likelihood problem with data $(\xbf_i$, $\zbf_i)$, and weights given by  by the approximate posterior, $q^{(t+1)}$.}
\end{itemize}
When the variational gap can be driven to zero, as in exact EM, this procedure is guaranteed to never decrease the likelihood in Equation \ref{eq:lik}. Moreover, the convergence properties of exact EM to both local and global minima have been carefully studied (\eg \citet{Dempster77, Neal1999, Balakrishnan2017}). Rigorous generalizations of EM to partial E- and M-steps also emerge naturally from this viewpoint \citep{Neal1999}.

One can also use the free energy viewpoint of EM to optimize a maximum \textit{a posteriori} (MAP) objective given by \\
\mbox{$\curlyL_{MAP}(\boldphi) \equiv \log p(\xbf|\boldphi) + \log p_0(\boldphi)$}, where $p_0$ is some prior distribution over parameters. This yields a corresponding free energy,
\begin{equation}\label{eq:fmap_def}
    F_{MAP}(q, \boldphi) \equiv \mathbb{E}_{q(\zbf)}[\log p(\xbf, \zbf|\boldphi) + \log p_0(\boldphi)] + H[q],
\end{equation}
upon which one can perform the same coordinate descent algorithm just described. This formulation is referred to as MAP-EM. We will draw connections between MAP-EM and EDAs with smoothed parameter updates such as those commonly used in CMA-ES. 

\section{Formal connection between EM and EDAs}
We are now ready to use the EM framework presented in Section \ref{sec:background}, to show that EDAs using the update rule in Equation \ref{eq:eda_obj} can be written as MC-EM, and as exact EM in the infinite sample limit. We then show that generalizations of Equation \ref{eq:eda_obj} that allow for parameter smoothing---such as CMA-ES's smoothing with the previous estimate and use of an evolutionary path---can be readily incorporated into the EM-EDA connection by using the MAP-EM framework. Finally, we highlight a connection between EM and standard gradient-based optimization.

\subsection{Derivation of EDAs as MC-EM}\label{sec:EDA-EM}
\ifthenelse{\boolean{FOOTNOTEHEADER}}{
\blfootnote{\footer}{}
}


As described in the introduction, EDAs seek to solve the objective defined in Equation \ref{eq:mbo}, 
\begin{align}
    \boldthetahat &\equiv
        \argmax_{\boldtheta} \expect_{p(\zbf|\boldtheta)}[f(\zbf)] \\
    &=
  \argmax_{\boldtheta} \log\expect_{p(\zbf|\boldtheta)}[f(\zbf)]\\
    &= \argmax_{\boldtheta} \leda(\boldtheta),
    \label{eq:EDAobj}
\end{align}
where $f(\zbf)$ is the black-box function to be optimized,  $p(\zbf|\boldtheta)$ is what we refer to as the {\it search model}, parameterized by $\boldtheta$, and we define \mbox{$\leda(\boldtheta) \equiv \log \expect_{p(\zbf|\boldtheta)}[f(\zbf)]$}. This expression can be thought of as an EDA equivalent to the `log marginal likelihood' in EM, only without any observed data, $\xbf$.

Some EDAs monotonically transform $f(\zbf)$ with a shaping function, $W(\cdot)$, which may be, for example, an exponential \citep{Peters2007}, a \\ quantile-based transformation \citep{Ollivier2017, Rubinstein1999}, or a cumulative density function (CDF) \citep{Brookes2019}. Although this transformation does not change the optima, it may alter the optimization dynamics. Often this shaping function is changed at each iteration in a sample-dependent, adaptive manner (which links these methods to annealed versions of EM and VI \citep{Hofmann2001, Mandt16}). In such a setting, the connection that we will show between EDAs and EM holds within each full iteration. For notational simplicity, we drop the $W(\cdot)$ and assume that $f(\zbf)$ has already been transformed. We additionally assume that the transformation is such that $f(\zbf) \geq 0$ for all $\zbf \in \curlyZ$.

To link Equation \ref{eq:EDAobj} to EM, we introduce a probability density, $q(\zbf)$, which allows us to derive a lower bound on $\leda$ using Jensen's inequality:
\begin{align}
    \leda (\boldtheta) &= \log\expect_{p(\zbf|\boldtheta)}[f(\zbf)] \label{eq:leda_def_1}\\
    &= \log\expect_{q(\zbf)}\left[\dfrac{p(\zbf|\boldtheta)f(\zbf)}{q(\zbf)}\right] \label{eq:leda_def_2}\\
    &\geq \expect_{q(\zbf)}\left[\log(  p(\zbf|\boldtheta) f(\zbf))\right] + H[q] \\
    &= F(q, \boldtheta), \label{eq:eda_fe}
\end{align}
where $F$ is the same free energy function appearing in the previous section on EM, except that the complete likelihood is replaced with the term $f(\zbf) p(\zbf|\boldtheta)$. When $f(\zbf) p(\zbf|\boldtheta)$ is normalizable,
then it can be shown that there is an EDA ``variational gap'' given by 
    \mbox{$F(q, \boldtheta) - \leda = -D_{KL}(q(\zbf)||\tilde{p}(\zbf|\boldtheta))$},
where we define the tilted density,
\begin{equation}
    \tilde{p}(\zbf|\boldtheta) = \dfrac{p(\zbf|\boldtheta)f(\zbf)}{\int_
    {\curlyZ} p(\zbf|\boldtheta)f(\zbf) d\zbf},
    \label{eq:tilted}
\end{equation}
which is the EDA counterpart to the exact posterior in EM. We can now construct a coordinate ascent algorithm on the free energy defined in Equation \ref{eq:eda_fe} that mirrors the EM algorithm. In particular, this algorithm iterates between E-steps that solve \\ \mbox{$q^{(t+1)} \gets \argmin_q D_{KL}(q(\zbf)||\tilde{p}(\zbf|\boldtheta^{(t)}))$}, and M-steps that solve \\ \mbox{$\theta^{(t+1)} \gets \argmax_\theta \expect_{q^{(t+1)}(\zbf)}\left[\log(  p(\zbf|\boldtheta) f(\zbf))\right]$}.
To make the precise connection between practically implemented EDA and EM, we introduce a particular approximate `posterior' for the E-step that is given by a mixture of weighted particles:
\begin{equation}
    \label{eq:approx_posterior}
    q^{(t+1)}(\zbf) = \dfrac{\sum_{i=1}^N f(\zbf_i) \delta_{\zbf_i} (\zbf)}{\sum_{i=1}^N f(\zbf_i)},
\end{equation}
where $\{\zbf_i\}_{i=1}^N$ are samples drawn from $p(\zbf|\boldtheta^{(t)})$, as in EDAs. Using this posterior approximation, the M-step amounts to solving the objective:
\begin{align}
        \boldtheta^{(t+1)} =& \argmax_{\boldtheta} \expect_{q^{(t+1)}(\zbf)}\left[\log (p(\zbf|\boldtheta)f(\zbf))\right] \label{eq:update_1}\\
        = &\argmax_{\boldtheta} \frac{\sum_{i=1}^N \int_{\curlyZ} f(\zbf_i)\delta_{\zbf_i}(\zbf) \log p(\zbf|\boldtheta)\, d\zbf}{\sum_{i=1}^N f(\zbf_i)} \\
        = &\argmax_{\boldtheta} \sum_{i=1}^N f(\zbf_i)\log p(\zbf_i|\boldtheta), \label{eq:EM-EDA}
\end{align}
which is exactly the generic EDA update step defined in Equation \ref{eq:eda_obj}. 

From this exposition it also becomes clear that EDAs can be thought of as performing a type of MC-EM that uses Importance Sampling~\cite{Mackay2003} rather than direct sampling from the posterior. In particular, the EDA sampling procedure uses proposal distribution, $p(\zbf|\theta^{(t)})$, (sampled in the E-step), and importance weights \\\mbox{$p(\zbf|\boldtheta)f(\zbf)/p(\zbf|\boldtheta)=f(\zbf)$} to estimate the expected ``complete log likelihood",  \mbox{$\expect_{q^{(t+1)}(\zbf)}\left[\log( p(\zbf|\boldtheta)f(\zbf))\right]$}, in the M-step.

This is our main result, as it shows that many EDAs can be viewed as an EM algorithm that uses the particle-based posterior approximation given by Equation \ref{eq:approx_posterior}.  
For any $\zbf$, we have
\\ \mbox{$q^{(t+1)}(\zbf) \overset{p}{\to} \tilde{p}(\zbf|\boldtheta^{(t)})$} as $n \to \infty$ by the law of large numbers and Slutsky's theorem \citep{grimmett2001probability}. In this limit of infinite particles, the approximate posterior matches the tilted distribution---the EDA equivalent to the ``exact posterior''---and our algorithm inherits the same guarantees as exact EM, such as guaranteed improvement of the objective function at each iteration, as well as local and global convergence properties \citep{Neal1999,Balakrishnan2017, Salakhutdinov2008}.


In many cases, EDAs use a member of an exponential family for the search model. Letting $\boldtheta$ denote the expectation parameters, then the search model has the density:
\begin{equation}\label{eq:exp_def}
    p(\zbf|\boldtheta) = h(\zbf)\exp\left(\boldeta(\boldtheta)^T T(\zbf) - A(\boldtheta)\right),
\end{equation} 
where $h$ is the base measure, $\boldeta(\boldtheta)$ are the natural parameters, $T(\zbf)$ are sufficient statistics and $A(\boldtheta)$ is the log-partition function. Then the update of Equation \ref{eq:EM-EDA} takes the simple form of a weighted maximum likelihood estimate for the exponential family:
\begin{equation}\label{eq:exp_update}
    \boldtheta^{(t+1)} = \dfrac{\sum_{i=1}^N f(\zbf_i)T(\zbf_i)}{\sum_{i=1}^N f(\zbf_i)}.
\end{equation}
Next we show how exponential family search models can be used to connect parameter-smoothed EDAs to MAP-EM.

\subsection{Smoothed EDAs as MAP-EM}

In CMA-ES, the parameter updates are smoothed between iterations in a number of ways~\citep{Hansen2003}. For example, the covariance estimate is typically smoothed with the previous estimate. Additionally, it may be further smoothed using a rank one covariance matrix obtained from the ``evolution path'' that the algorithm has recently taken.\footnote{See the two terms in Equation 11 in \citet{Hansen2003}).} Next, we consider these types of smoothing in detail. However, any smoothing update, or combination thereof, can be similarly derived by adjusting the form of the prior distribution. A benefit of viewing CMA-ES updates in this general form is that it becomes more straightforward to determine how to do similar types of smoothing for EDAs without Gaussian search models.

When smoothing with the previous parameter estimate, the updates can be written as
\begin{equation}\label{eq:smoothed_update}
    \boldtheta^{(t+1)} = (1-\gamma) \boldtheta^{(t)} + \gamma \tilde{\boldtheta}^{(t+1)},
\end{equation}
where $\tilde{\boldtheta}^{(t+1)}$ is the solution to Equation \ref{eq:EM-EDA} and $\gamma$ is a hyperparameter that controls the amount of smoothing. In the case where the search model is a member of an exponential family, as is defined in Equation \ref{eq:exp_def}, we will show that the smoothed update of Equation \ref{eq:smoothed_update} is equivalent to a particular MAP-EM update that uses the tilted density of Equation \ref{eq:tilted} as the approximate posterior. To see this, consider the conjugate prior to the exponential family,
\begin{equation}\label{eq:priordef}
    p_0(\boldtheta|\boldlambda) = \exp\left(\boldlambda_1^T \boldeta(\boldtheta) - \lambda_2 A(\boldtheta) - B(\boldlambda) \right),
\end{equation}
where $\boldlambda = (\boldlambda_1, \lambda_2)$, $B(\boldlambda)$ is the log-partition function of the prior, and we have assumed that the base measure is constant. We now consider the modified EDA objective: $$\hat{\curlyL}_{EDA} (\boldtheta) = \log\expect_{p(\zbf|\boldtheta)}[f(\zbf)p_0(\boldtheta|\boldlambda)],$$ which is analogous to the MAP objective defined in Section \ref{sec:background}. It can be shown that by replacing Equation \ref{eq:leda_def_1} with this modified objective, performing analogous steps as those in Equations (\ref{eq:leda_def_2}-\ref{eq:eda_fe}) and (\ref{eq:update_1}-\ref{eq:EM-EDA}), and using the same definition of the tilted density and approximate posterior as in Equations \ref{eq:tilted} and \ref{eq:approx_posterior}, respectively, we arrive at the update equation:
\begin{equation}\label{eq:map_update}
    \boldtheta^{(t+1)} = \argmax_{\boldtheta} \sum_{i=1}^N f(\zbf_i) \log p(\zbf_i|\boldtheta)p_0(\boldtheta|\boldlambda).
\end{equation}
If we now let $\lambda_2 = \nicefrac{1}{\gamma} - 1$ and allow $\boldlambda_1$ to change every iteration as $\boldlambda_1 = (\nicefrac{1}{\gamma} - 1)\boldtheta^{(t)}$, it can then be shown that the solution to this objective is given by:
\begin{align}\label{eq:map_em_update}
\boldtheta^{(t+1)} &= \dfrac{\boldlambda_1}{1+\lambda_2}
\boldtheta^{(t)}
+ \dfrac{1}{1+\lambda_2} \tilde{\boldtheta}^{(t+1)} \\
&= (1-\gamma) \boldtheta^{(t)} + \gamma \tilde{\boldtheta}^{(t+1)}, \label{eq:smooth2}
\end{align}
where $\tilde{\boldtheta}^{(t+1)} = \frac{\sum_{i=1}^N f(\zbf_i)T(\zbf_i)}{\sum_{i=1}^N f(\zbf_i)}$ is Equation \ref{eq:exp_update}, the solution of original EDA objective. Since the update in Equation \ref{eq:smooth2} is identical to the smoothed update of Equation \ref{eq:smoothed_update},  we can see that smoothing can be viewed as a consequence of performing MAP-EM in the context of EDAs, with a particular choice of prior parameters. See the Appendix \ref{map_update_app} for a full derivation.

\subsection{A continuum between EM-like algorithms and gradient descent}

\label{sec:gradient}

Instead of using EDAs, one could alternatively solve the objective in Equation \ref{eq:mbo} directly with gradient descent (or natural gradient descent) where updates are given by
    \mbox{$\boldtheta^{(t+1)} = \boldtheta^{(t)} + \alpha \nabla \leda(\boldtheta^{(t)})$},
and $\alpha$ is a step size parameter. This is known as REINFORCE \citep{Williams92} in the RL community and is related to IGO-ML \citep{Ollivier2017}. Typically, one cannot compute the gradient term exactly, in which case the ``log derivative trick'',
\begin{align}
    \nabla_{\boldtheta} &\expect_{p(\zbf|\boldtheta)}[f(\zbf)] |_{\boldtheta=\boldtheta^{(t)}} \\ &= \expect_{p(\zbf|\boldtheta^{(t)})}[f(\zbf)\nabla_{\boldtheta} \log p(\zbf | \boldtheta)|_{\boldtheta=\boldtheta^{(t)}}],
\end{align}
    is often combined with Monte Carlo estimation to arrive at an update:
\begin{equation}\label{eq:sgd_update}
    \boldtheta^{(t+1)} = \boldtheta^{(t)} + \alpha \sum_{i=1}^N f(\zbf_i)\nabla_\theta\log p(\zbf_i|\boldtheta)|_{\boldtheta=\boldtheta^{(t)}},
\end{equation}
 where $\{\zbf_i\}_{i=1}^N$ are samples drawn from $p(\zbf|\boldtheta^{(t)})$. Notably, this update does not require the gradient of $f(\zbf)$ \citep{Recht2018}.
 %

%
We can connect the gradient-based optimization in Equation \ref{eq:sgd_update} to the EDA ``EM'' optimization presented in Section \ref{sec:EDA-EM}. In particular, consider the partial M-step version of EM, where instead of fully solving for the new parameter, one instead performs one gradient step ---so called ``first-order'' EM \citep{Neal1999, Balakrishnan2017}. Instantiating first-order EM into EDA ``EM'' by partially solving the objective in Equation \ref{eq:EM-EDA} with a single gradient step (with step size $\alpha$) will result in the same update as Equation \ref{eq:sgd_update}. In other words, performing ``first-order" EM in the EDA context is identical to performing gradient descent on Equation \ref{eq:mbo}. As we can see, as one performs more gradient steps within an M-step, one traces out a continuum of methods between gradient descent on the original objective and EM.

\section{A unifying view of EDAs as natural gradient descent}

In the next sections, we provide an illustrative example of how the connection between EDAs and EM can be leveraged. In particular, we use the framework to provide a new perspective on how EDAs can be seen as approximating natural gradient descent.
 
\citet{Akimoto2010} showed that CMA-ES, a particular EDA, can be viewed as approximate natural gradient descent (NGD) . This result also emerges from the IGO framework of \citet{Ollivier2017}, who additionally showed that CEM and several other EDAs can be viewed as NGD. \citet{Malago2013} also show how EDAs using MRF for the search model can be interpreted as NGD.
  The relationship we derived between EDA and EM provides a unifying perspective on these connections, which also allows for them to be readily generalized to any EDA that can be written as EM. Next we develop this unifying view.
\ifthenelse{\boolean{FOOTNOTEHEADER}}{
\blfootnote{\footer}{}
}

  It does not appear widely known that EM can be viewed as approximate NGD. Thus we next synthesize a series of known, but independent, results to show this.
  As a method for maximizing the log-likelihood function, Chr\'etien and Hero showed that EM can be formulated as the proximal point method (PPM) with the reverse KL divergence \citep{chretien98, chretien2000kullback}. Specifically, this formulation shows that in our context, each update
\begin{align}
    \boldtheta^{(t + 1)} = &\argmax_{\boldtheta} \int_{\curlyZ} \tilde{p}(\zbf | \boldtheta^{(t)}) \log (p(\zbf|\boldtheta)f(\zbf))\, d \zbf
\end{align}
is equivalent to
\begin{align}
     \boldtheta^{(t + 1)} = \argmax_{\boldtheta} \left[\leda(\boldtheta) - D_{KL}(\tilde{p}(\zbf| \boldtheta^{(t)}) || \tilde{p}(\zbf| \boldtheta)) \right],
\end{align}
which is a PPM (see also Appendix \ref{ppm_app}). This equivalence requires the exact posterior (tilted distribution), $\tilde{p}$, and therefore does not technically hold for EDA in practice, which is more akin to MC-EM. Nevertheless, the connection exposes an interesting new perspective on EDAs in the asymptotic regime, as we shall see.

\cut{
The use of ``proximal penalties" at each iteration of the PPM can beneficially stabilize the updates in iterative algorithms \citep{chretien2000kullback}. 
 Additionally, the PPM is particularly important in non-smooth optimization because proximal operators define a smooth approximation of the objective function\cut{, \mbox{$y \to \sup_{x} f(x) + \frac{1}{\gamma} D(y, x) $},} known as the Bregman-Moreau envelope \citep{parikh2014proximal,chen2012moreau,bauschke2018regularizing}; thus the PPM can be viewed as gradient descent on a smooth Bregman-Moreau envelope. In the special case that the divergence is 
  squared-Euclidean distance, the smoothed objective is known as the Moreau envelope and the PPM is also known as Moreau-Yosida regularization \citep{parikh2014proximal}, whereas here, the use of the KL divergence, the smoothing is attuned to the underlying geometry, which as we shall see, connects us to NGD.
 ensures that the smoothing reflect the underlying geometry.}

Next, consider the mirror descent update, which is a linearization of the PPM \citep{beck2003mirror}:
 \begin{align}
    \boldtheta^{(t + 1)} & = \argmax_{\boldphi} \leda(\boldtheta^{(t)}) \\ & +  \langle \nabla_{\boldtheta} \leda(\boldtheta) \big|_{\boldtheta = \boldtheta^{(t)}}, \boldtheta - \boldtheta^{(t)} \rangle \\ & - D_{KL}(\tilde{p}(\zbf| \boldtheta^{(t)}) || \tilde{p}(\zbf|\boldtheta)).
\end{align}
Mirror descent with KL divergence is equivalent to NGD on the dual Reimannian manifold \citep{raskutti2015information}. Even when the original and dual manifolds differ, the two algorithms can be related: using a second-order Taylor series approximation of the KL divergence yields an approximate mirror descent update that is precisely NGD. Specifically, replacing the KL divergence in mirror descent with the second-order approximation as follows,
\begin{align}
    D_{KL}&(\tilde{p}(\zbf| \boldtheta^{(t)}) || \tilde{p}(\zbf| \boldtheta)) \\
    & = \frac{1}{2} (\boldtheta - \boldtheta^{(t)})^\top I(\boldtheta^{(t)})  (\boldtheta - \boldtheta^{(t)}) + O\left( (\boldtheta - \boldtheta^{(t)})^3 \right),
\end{align}
where $I$ is the Fisher information metric~\citep{martens2014new}, yields the update
\begin{align}
    \boldtheta^{(t + 1)} & = \argmax_{\boldtheta} \leda(\boldtheta^{(t)}) \\
    & + \langle \nabla_{\boldtheta} \leda(\boldtheta) \big|_{\boldtheta = \boldtheta^{(t)}}, \boldtheta - \boldtheta^{(t)} \rangle \\
    & - \frac{1}{2} (\boldtheta - \boldtheta^{(t)})^\top I(\boldtheta^{(t)})  (\boldtheta - \boldtheta^{(t)}),
\end{align}
which can be shown to be equivalent to the NGD update 
\ifthenelse{\boolean{ARXIV}}
{\begin{align}
    \boldtheta^{(t + 1)} =  \boldtheta^{(t)} + I(\boldtheta^{(t)})^{-1} \nabla_{\boldtheta} \leda(\boldtheta) \big|_{\boldtheta = \boldtheta^{(t)}}.
\end{align}
}
{
    \mbox{$\boldphi^{(t + 1)} =  \boldphi^{(t)} + I(\boldphi^{(t)})^{-1} \nabla_{\boldphi} \log \mathcal{L}(\boldphi) \big|_{\boldphi = \boldphi^{(t)}}$}.
}
 See Appendix \ref{si:NGD} for further details.
\\ \\

\noindent{\bf In summary:} EM is a PPM with KL divergence; mirror descent is a linearization of the PPM; and a second-order approximation of the KL divergence in mirror descent yields NGD. Thus in the infinite sample limit, EDAs can be seen as approximate NGD.

Note that \citet{Sato2001}, and later others, showed the related result that for exponential family models with hidden variables and a conjugate prior (to the joint observed and hidden model), classical (non-amortized) mean-field variational inference performs NGD \citep{ Hensman2012, Hoffman2013}. In contrast, our result applies to any exact EM, but only connects it to approximate NGD.

An interesting side note is that \citet{Salimbeni2018}~demonstrated empirically that the best step size for NGD, according to Brent line searches, increases over iterations to 1.0, for Gaussian approximations of several models. Their result aligns nicely with our results here that EM, viewed as approximate NGD, has a step size of precisely 1.0.

\section{Discussion}

We have shown a novel connection between a broad class of EDAs and EM. Additionally, we have presented an illustrative example of insight from EM that can be applied to EDAs by way of this connection. Specifically, we presented a new connection between NGD and EDAs, by way of EM.
 
 In light of EDAs as EM, one can ask whether this viewpoint sheds any light on choices of the EDA search model that may be more or less optimal. Suppose for a moment that rather than a sample-based approximate posterior, one instead used a parametric posterior, as in standard VI. In such a case, if the search model were in some sense ``conjugate" to $f(\zbf)$, then the posterior could take on an analytical form. When might such an approach make sense? Suppose one were using EDAs to perform protein design \citep{Brookes2019}, that $f(\zbf)$ represented protein stability, and that $f(\zbf)$ could be approximated by using an exponential model of linear additive marginal and pairwise terms \citep{Bryngelson7524}. Then if one used a Potts model \citep{Marks2011} as the search model, the exact posterior could also be written in this form, and a ``parametric" EDA could be pursued. 
  
  \cut{
EDAs are often improved by smoothing, such as exponentially decaying averaging of previous updates, or model-specific schemes such as in CMA-ES \citep{Hansen2001}. We did not employ such smoothing here, but expect that doing so would not alter the message of our paper.
}

Given the tight connection between EDAs with adaptive shape functions, and annealed versions of EM and VI \citep{Hofmann2001, Mandt16}, we believe that these latter approaches could benefit from the simple and robust implicit annealing schemes found in the EDA literature, which arise not from an annealing schedule, but from simple and effective quantile transformations and the like.







\appendix
\section{Derivations in the case of Exponential Family search models}

In Equations \ref{eq:exp_update} and \ref{eq:map_em_update} we present exact update equations for EDAs when the search model is a member of an exponential family. We derive these updates here.

\subsection{EDA Update}\label{eda_update_app}
For the in update, Equation \ref{eq:exp_update}, we use the definition of the exponential family search model density in Equation \ref{eq:exp_def} in the EDA update objective in Equation \ref{eq:EM-EDA}:
\begin{align}
    \boldtheta^{(t+1)} &= \argmax_{\boldtheta} \sum_{i=1}^N f(\zbf_i)\log p(\zbf_i|\boldtheta) \\
    &= \argmax_{\boldtheta} \left(\sum_{i=1}^N f(\zbf_i) \boldeta(\boldtheta)^T T(\zbf_i)\right) + A(\boldtheta) \sum_{i=1}^N f(\zbf_i) 
\end{align}
To solve this, we now take the gradient of the argument with respect to $\boldtheta$:
\begin{align}
   & \nabla_{\boldtheta} \left(\sum_{i=1}^N f(\zbf_i) \boldeta(\boldtheta)^T T(\zbf_i)\right) + \nabla_{\boldtheta} A(\boldtheta) \sum_{i=1}^N f(\zbf_i) \\
   = &\nabla_{\boldtheta} \boldeta(\boldtheta)^T \sum_{i=1}^N f(\zbf_i) T(\zbf_i) + \nabla_{\boldtheta} \boldeta(\boldtheta)^T \expect_{\boldeta}[T(\zbf)] \sum_{i=1}^N f(\zbf_i) \label{eq:grad_exp_fam1}
\end{align}
where we have used the chain rule and the properties of the log-partition function (i.e. cumulant generating function) to equate $\nabla_{\boldtheta} A(\boldtheta) = \nabla_{\boldtheta} \boldeta(\boldtheta) \expect_{\boldeta}[T(\zbf)]$. Now, recognizing that $\boldtheta$ are expectation parameters, so $\boldtheta = \expect_{\boldeta}[T(\zbf)]$, and setting Equation \ref{eq:grad_exp_fam1} equal to zero, we arrive at the update in  Equation \ref{eq:exp_update}:
\begin{align}
    \nabla_{\boldtheta} \boldeta(\boldtheta)^T \sum_{i=1}^N f(\zbf_i) T(\zbf_i) &+ \nabla_{\boldtheta} \boldeta(\boldtheta)^T \expect_{\boldeta}[T(\zbf)] \sum_{i=1}^N f(\zbf_i) \label{eq:grad_exp_fam2} = 0 \\
    \Rightarrow \quad \quad \boldtheta^{(t+1)} &= \dfrac{\sum_{i=1}^N f(\zbf_i)T(\zbf_i)}{\sum_{i=1}^N f(\zbf_i)}
\end{align}

\subsection{MAP Update}\label{map_update_app}
\ifthenelse{\boolean{FOOTNOTEHEADER}}{
\blfootnote{\footer}{}
}

For the update in Equation \ref{eq:map_em_update}, we use the definition of the exponential family search model density in Equation \ref{eq:exp_def}, and the prior in Equation \ref{eq:priordef}, in the MAP-EDA update objective, Equation \ref{eq:map_update}:
\begin{align*}
    \boldtheta^{(t+1)} &= \argmax_{\boldtheta} \sum_{i=1}^N f(\zbf_i) \log p(\zbf_i|\boldtheta)p_0(\boldtheta|\boldlambda) \\
    &= \argmax_{\boldtheta} \sum_{i=1}^N f(\zbf_i) \boldeta(\boldtheta)^T T(\zbf_i) + \left(\boldlambda_1^T \boldeta(\boldtheta) +(1+\lambda_2) A(\boldtheta)\right)\sum_{i=1}^N f(\zbf_i)
\end{align*}

Taking the gradient of the argument with respect to $\boldtheta$ and setting it equal to zero:

\begin{align*}
    &\nabla_{\boldtheta} \boldeta(\boldtheta)^T \sum_{i=1}^N f(\zbf_i)  T(\zbf_i) \\ &\qquad + \left(\nabla_{\boldtheta} \boldeta(\boldtheta)^T \boldlambda_1 + (1+\lambda_2) \nabla_{\boldtheta} \boldeta(\boldtheta)^T  \expect_{\boldeta}[T(\zbf)]\right)\sum_{i=1}^N f(\zbf_i) = 0 \\
    &\Rightarrow \quad \quad \boldtheta (1+\lambda_2) \sum_{i=1}^N f(\zbf_i) = \sum_{i=1}^N f(\zbf_i)  T(\zbf_i)  + \boldlambda_1 \sum_{i=1}^N f(\zbf_i) \\
    &\Rightarrow \quad \quad \boldtheta^{(t+1)} = \dfrac{1}{1+\lambda_2}\dfrac{\sum_{i=1}^N f(\zbf_i)T(\zbf_i)}{\sum_{i=1}^N f(\zbf_i)} + \dfrac{\boldlambda_1}{1+\lambda_2}
\end{align*}
where again we use $\boldtheta = \expect_{\boldeta}[T(\zbf)]$. Now, we can recognize $\tilde{\boldtheta}^{(t+1)} = \frac{\sum_{i=1}^N f(\zbf_i)T(\zbf_i)}{\sum_{i=1}^N f(\zbf_i)}$ is the solution to the original EDA objective, and letting $\lambda_2 = \nicefrac{1}{\gamma} - 1$ and $\boldlambda_1 = (\nicefrac{1}{\gamma} - 1)\boldtheta^{(t)}$ we arrive at:

\begin{align*}
    \boldtheta^{(t+1)} &= \dfrac{1}{1+\nicefrac{1}{\gamma} - 1}\tilde{\boldtheta}^{(t+1)} + \dfrac{\nicefrac{1}{\gamma} - 1}{1+\nicefrac{1}{\gamma} - 1} \boldtheta^{(t)}\\
    &= \gamma \tilde{\boldtheta}^{(t+1)} + (1-\gamma) \boldtheta^{(t)}
\end{align*}
which is the desired smoothed update.

\section{Equivalence between Expectation-Maximization and the proximal point method}\label{ppm_app}

In Section 6, we discuss the equivalence between EM and using the proximal point method (PPM) with reverse KL divergence to maximize the log-likelihood function \cite{chretien98, chretien2000kullback}. Here, we provide the derivation of this equivalence, beginning with a formulation of the PPM:
\begin{align*}
    \boldphi^{(t + 1)} &= \argmax_{\boldphi} \log \mathcal{L}(\boldphi) - D_{KL}(p(\zbf| \ybf, \boldphi^{(t)}) || p(\zbf| \ybf, \boldphi))\\
    &= \argmax_{\boldphi} \log p(\ybf | \boldphi) - \int p(z| \ybf, \boldphi^{(t)}) \log \frac{p(z| \ybf, \boldphi^{(t)})}{p(z| \ybf, \boldphi)} dz \\
    &= \argmax_{\boldphi} \log p(\ybf | \boldphi) + \int p(z| \ybf, \boldphi^{(t)}) \log p(z| \ybf, \boldphi) dz \\
    &= \argmax_{\boldphi} \log p(\ybf | \boldphi) \int p(z| \ybf, \boldphi^{(t)}) dz \\ &\qquad + \int p(z| \ybf, \boldphi^{(t)}) \log p(z| \ybf, \boldphi) dz \\
    &= \argmax_{\boldphi} \int p(z| \ybf, \boldphi^{(t)}) \left( \log p(\ybf | \boldphi) + \log p(z| \ybf, \boldphi) \right) dz \\
    &= \argmax_{\boldphi} \int p(z| \ybf, \boldphi^{(t)}) \log p( \ybf, z | \boldphi) dz. 
\end{align*}
Note that the last line is precisely the update rule for EM.

\section{Details of natural gradient descent}
\label{si:NGD}

Recall from Section 6 that replacing the log-likelihood with its linearization and the KL-divergence with its second-order Taylor series approximation in the PPM formulation of EM yields the update
\begin{align*}
    \boldphi^{(t + 1)} &= \argmax_{\boldphi} \log \mathcal{L}(\boldphi^{(t)}) + \langle \nabla_{\boldphi} \log \mathcal{L}(\boldphi) \big|_{\boldphi = \boldphi^{(t)}}, \boldphi - \boldphi^{(t)} \rangle \\ &\qquad - \frac{1}{2} (\boldphi - \boldphi^{(t)})^\top I(\boldphi^{(t)})  (\boldphi - \boldphi^{(t)}).
\end{align*}
One can further derive a closed-form update by taking the derivative of the $\argmax$ argument on the right-hand side, setting it equal to zero as per first-order optimality conditions, and solving:
\begin{align*}
    &\nabla_{\boldphi} \big( \log \mathcal{L}(\boldphi^{(t)}) + \langle \nabla_{\boldphi} \log \mathcal{L}(\boldphi) \big|_{\boldphi = \boldphi^{(t)}}, \boldphi - \boldphi^{(t)} \rangle \\ &\qquad - \frac{1}{2} (\boldphi - \boldphi^{(t)})^\top I(\boldphi^{(t)})  (\boldphi - \boldphi^{(t)}) \big) \big|_{\boldphi = \boldphi^{(t + 1)}} = 0 \\
    & \implies \nabla_{\boldphi} \log \mathcal{L}(\boldphi) \big|_{\boldphi = \boldphi^{(t)}} - I(\boldphi^{(t)}) (\boldphi^{(t + 1)} - \boldphi^{(t)}) = 0 \\
    & \implies \boldphi^{(t + 1)} =   \boldphi^{(t)} + I(\boldphi^{(t)})^{-1} \nabla_{\boldphi} \log \mathcal{L}(\boldphi) \big|_{\boldphi = \boldphi^{(t)}}.
\end{align*}
This establishes the equivalence between the standard natural gradient descent update rule and the mirror descent update derived using the PPM formulation of EM as a starting point.


\begin{acks}
We thank Ben Recht for helpful discussion, David Duvenaud, Roger Grosse and James Hensman for pointers to variational Bayes as NGD, Petros Koumoutsakos for encouraging us to think about how to incorporate smoothing, Nikolaus Hansen for pointing us to relevant work, Youhei Akimoto for replies to questions about his papers, and Meng Xiangming for catching technical typos.

This research used resources of the National Energy Research
Scientific Computing Center, a DOE Office of Science User Facility
supported by the Office of Science of the U.S. Department of Energy
under Contract No. DE-AC02-05CH11231.
\end{acks}

\bibliographystyle{ACM-Reference-Format}
\bibliography{main} 

\end{document}